\documentclass[runningheads]{llncs}
\usepackage[T1]{fontenc}
\usepackage{graphicx}
\usepackage{comment}
\begin{document}
\title{Ontological analysis of proactive life event services}
\author{Kuldar Taveter\orcidID{0000-0003-3074-7618}}
\authorrunning{K. Taveter}
\institute{Institute of Computer Science, University of Tartu, Tartu, Estonia 
\email{kuldar.taveter@ut.ee}\\
}
\maketitle              % typeset the header of the contribution
\begin{abstract}
Life event service is a direct digital public service provided jointly by several governmental institutions so that a person can fulfill all the obligations and use all the rights that arise due to a particular event or situation in personal life. Life event service consolidates several public services related to the same life event into one service for the service consumer. This paper presents an ontological analysis of life event services, which is based on the works by Guarino, Guizzardi, Nardi, Wagner, and others. The purpose of the ontological analysis is to understand the meanings of life event, proactive public service based on life event, and other related notions. This kind of ontological analysis is crucial because for implementing the hardware and software architectures of e-government and digital public services, it is essential to agree upon the precise meanings of the underlying terms.

\end{abstract}
\section{Introduction}
%\subsection{A Subsection Sample}
%Please note that the first paragraph of a section or subsection is
%not indented. The first paragraph that follows a table, Fig.,
%equation etc. does not need an indent, either.

%Subsequent paragraphs, however, are indented.

The Government of Estonia has stated in the regulation “Principles for Managing Services and Governing Information” \cite{ref_url1} a new direction towards “invisible” public services which are combined around a person’s life event in a manner that they seem as one comprehensive and smooth service. The life event services should require no more than one interaction between a person and an institution, moreover, ideally are delivered fully automatically without any interactions.
The Estonian digital society development plan for the year 2030 \cite{ref_url2} describes the next development leap of the digital country with public services that are event-based, proactive, human-centered and powered by Artificial Intelligence (AI). The development of digital services must be based on technological and social trends and the changed expectations and needs of society. The next stage of development of the digital government will link these challenges and opportunities into a complete personal government. A personal government \cite{ref_url3} means redesigning public services so that the complexity of the public sector remains in the background for their consumers. They get their actions carried out as simply as possible and when necessary. With the same meaning, the term “algorithmic government” has been used in the research literature \cite{ref_proc7}.  

One of the cornerstones of personal government is the concept of life event service. Life event service is a direct digital public service provided jointly by several institutions so that a person can fulfill all the duties and use all the rights that arise due to a particular event or situation in personal life \cite{ref_url1}. Life event service consolidates several public services related to the same life event into one service for the service consumer. The concept of life event services creates a comprehensive picture of the public services for a person, including the person's duties and rights. The design and implementation of life event services requires a proper understanding and agreeing about the relevant concepts and relationships between them. This kind of understanding can be obtained through a proper ontological analysis of the problem domain.

This paper presents an ontological analysis \cite{ref_book1} of life event services. The purpose of the ontological analysis is to understand the meanings of life events, proactive public service based on life events, and other related notions. This kind of ontological analysis is crucial because for implementing the hardware and software architectures of e-government and digital public services, it is essential to agree upon the precise meanings of the underlying terms. The paper presents and further develops the ontological analysis of life event services performed by the author in the project “Life event services analysis”, which was initiated by the Estonian Ministry of Economic Affairs and Communication and performed in February 2019 – May 2020 \cite{ref_url4}.

The rest of this paper is structured as follows. Section \ref{background} describes the ontological concepts and relationships forming the foundation of the ontological metamodel of life event services. Thereafter, Section \ref{analysis} presents an ontological analysis of proactive life event services. Finally, Section \ref{conclusions} draws conclusions and outlines future work.

\section{Background of proactive life event services}
\label{background}

Ontological analysis \cite{ref_book1} divides entities into universals and particulars. Universals are abstract concepts that do not exist in time and space - that is, they cannot be localized. Examples of universals are mathematical objects such as numbers and sets, modeling constructs such as goals and roles, and types such as types of agents, objects, and activities. Universals are patterns or properties that materialize in many different particulars. 

\begin{figure}
    \centering
    \includegraphics{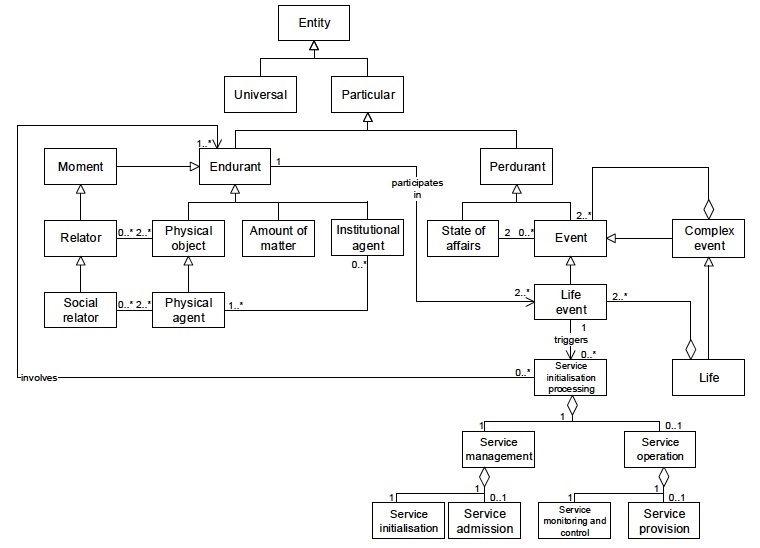}
    \caption{Ontological metamodel of life event services}
    \label{fig:metamodel}
\end{figure}

Particulars, on the other hand, are entities that exist at least in time. An example of an entity that exists in time but not necessarily in space is a computer program. Each particle has a specific identity. Particulars can be distinguished from universals by the ``causality criterion'', according to which a particular can cause another particular, while there is no such causal connection between universals. In the context of life events and event services, we are more interested in particulars, which is why we discuss them in more detail below. The division of entities into universals and particulars is expressed in Fig. \ref{fig:metamodel}, which depicts the ontological metamodel of life event services.

Ontologically, two types of particulars are distinguished: endurants and perdurants. They are distinguished by their behavior over time. Simply put, while endurants ``exist in time'', perdurants ``happen in time''. An endurant is accordingly defined as an entity that maintains its identity over time. Examples of an endurant are a house, the moon, and a pile of sand. Unlike the endurant, if the perdurant exists, not all of its parts exist at the same time. Examples of a perdurant include humanity, conversation, World War II, and a business process. One can intuitively distinguish between endurants and perdurants as between things and processes. For example, this paper is an endurant because it exists as a whole, while “your reading of the paper” is perdurant because at this point you are no longer reading the previous paragraph and you are not yet reading the next paragraph. The distribution of particulars into endurants and perdurants is represented by Fig. \ref{fig:metamodel}.

An endurant can be a physical object, some amount of matter, or an institutional agent. A physical object, such as a house, a person, the Moon or a computer program is a type of endurant that satisfies the condition of unity, meaning that certain parts of this endurant can change without changing the identity of the endurant. In contrast, an amount of matter, such as a pile of sand, does not satisfy the condition of unity, so when any part of such an endurant changes, the identity of the endurant generally changes as well. A physical agent such as a person or computer program containing AI is a type of physical object that can perceive its environment, perform logical reasoning, and perform actions. An institutional agent is a set consisting of physical agents located within it, who share collective knowledge and perceive its environment on behalf of the institutional agent, and perform logical inferences and perform actions on behalf of the institutional agent. Endurant subtypes are presented in Fig. \ref{fig:metamodel}.

Endurants are characterized by moments. Moments are themselves endurants, which are existentially dependent on other endurants, which are called their bearers. Moments are divided into two: internal moments and relator moments. An intrinsic moment is such a moment that depends existentially on one specific carrier. For example, the color of the apple in one’s hand depends on the existence of this particular apple. A relator is a moment that depends existentially on more than one physical object, such as a flight connection between two different cities. A social relator is a relator that connects two or more physical agents. A social relator is, for example, marriage. Fig. \ref{fig:metamodel} expresses the concepts of moment, relator and social relator.

The two main types of perdurant are state and event. The state can be explained through perdurant's imaginable elementary time parts - atomic time moments or snapshots. The state of an entity means that a given entity can be characterized in the same way at successive atomic moments of time. For example, the state of John can be described with the expression ``John is sitting on a chair''.

An event, on the other hand, is a perdurant, which is related to two states - the collective state of the world's entities before and after the event. Such collective state of the entities of the world is termed as state of affairs. For example, the expression ``John got up from the chair'' describes an event that separates the state of the world where John was sitting on the chair from the state of the world where John is no longer sitting on the chair. The concepts of event and state of affairs are represented in Fig. \ref{fig:metamodel}.

It is important to distinguish between atomic events and complex events. An atomic event is considered as an event that takes no time to occur, such as an explosion or receiving a message by a mobile phone. Complex events, such as a storm, a birthday party or the Second World War, consist of many other events. A person's life is a complex event that connects all the events that take place in a person's life span, while many events - for example, pregnancy and the birth of a child - have causal relationships, which can be utilized by proactive services based on life events.

For the purposes of the topic of this paper, the most important universals are types, roles, and phases. A type is a universal that defines a set of particulars, each member of which has a specific identity and each member of which maintains this identity in every situation that can be represented by the given ontological model. For example, Human is a type because it carries a specific identity that determines who can be defined as human. In the same way, Citizen is also a type, which carries a specific identity, on the basis of which it can be decided who can be considered a Citizen within the given ontological model, which may or may not coincide with, for example, the concept of a Citizen of the Republic of Estonia. Also important is the fact that the ontological model must treat each Citizen as an endurant - a physical agent - with one and the same specific identity throughout the endurant's whole lifetime. This means that the identity of a specific person cannot legally change in state registers, databases and information systems.

Types are divided into species, subspecies, phases and roles. A species is a strict type, meaning that any endurant of a given species is inevitably a member of that species - it cannot change its type. Examples include the species Human and Dog, of which endurants cannot change species, except in fairy tales and science fiction.

Species can be divided into their strict subtypes - subspecies. The subspecies inherits the principle of determining the identity of the endurants belonging to it from the given species, but specifies this principle so that it only applies to some of the endurants belonging to the given species. For example, the species Human is divided into the subspecies Male and Female, the endurants of which as a general rule cannot change their subtype, except perhaps through a gender-change operation.

In addition to strict types - species and subspecies - there are also non-strict types: roles and phases. A role is a non-strict type, as endurants belonging to it can change their type. For example, a person fulfilling the role of Citizen can give up this role by taking the citizenship of another country and/or by moving away from Estonia. A role is defined through the duties (responsibilities) that endurants fulfilling a given role must fulfill, as well as through its rights. An example of duties is the duty to complete military service in the sub-role Conscript of the role Citizen. An example of rights is the right to use one or other e-government service. In certain situations, it is also necessary to point out the constraints of the role, which determine explicitly what the person performing the given role must not do. An example of a constraint is the constraint set by the Marriage Act that the persons performing the roles of Wife and Husband forming a family cannot be close relatives of each other. The duties, rights and constraints of an institutional agent are performed on behalf of the respective institutional agent by its internal physical agents - people. For example, every organization has one or more signatory representatives of that organization who have the right to assume duties, exercise rights and set constraints on behalf of the given organization. Duties, rights and constraints of an organization are also known as deontic assignments \cite{ref_proc1}.

Phases, also called states, are non-strict types that represent the evolution of endurants of a given type over time. For example, the subspecies Male has phases Boy, Teenager, and Adult Male.

%A goal is ontologically defined as a state of the world desired by one or more physical agents. For example, the goal may be to get married or have a child or finish school or drive a car. An objective can be divided into optional or mandatory sub-objectives. As an example of optional goals, the goal of having a child can be achieved through childbirth or adoption. An example of mandatory goals is a situation where both a theory and a driving test must be passed in order to drive a car. A subtype of goal is intention, which can be defined as a goal that one or more agents have taken to fulfill, and for the fulfillment of which the agents have a plan specifying specific actions and the order in which they are performed.

\section{Analysis of proactive life event services}
\label{analysis}

Life event depicted in Fig. \ref{fig:metamodel} is an event that separates two different states of at least one physical or institutional agent. As we explained in Section \ref{background}, such states are termed as states of affairs. Examples of life events are marriage, pregnancy, birth of a child, creation of a company, naming a child, expansion of the company's share capital, obtaining management rights, property acquisition and retirement.

A life event can be reflected by \textit{data events} in various state registers, information systems and databases, and based on this, various activities related to the provision of public services can be triggered.

A process is a complex event consisting of two or more sequential or parallel events. Examples of a process are a storm, a football game, a conversation, a birthday party, and shopping in an online store. Activity is a process performed by one or more agents. For example, running is an action because John runs, but boiling is a chemical process that does not involve any agent performing it. Activity consists of atomic actions. An activity in which two or more agents participate includes interactions between these agents.

Public service is a set of all activities that realize the obligation of a public institutional agent to make available to individuals - physical agents - or other institutional agents - organizations - capabilities that meet their needs, also giving them opportunities to control how and when such capabilities can be used \cite{ref_proc2}. According to \cite{ref_article1}, ability is ontologically defined as the property of a physical or institutional agent to perform certain actions or create certain outputs under certain conditions. In the case of public services, such specific conditions are, for example, the Citizen's right to consume the corresponding service and the availability of the resources necessary to consume the service.

\begin{figure}
    \centering
    \includegraphics[width=\textwidth]{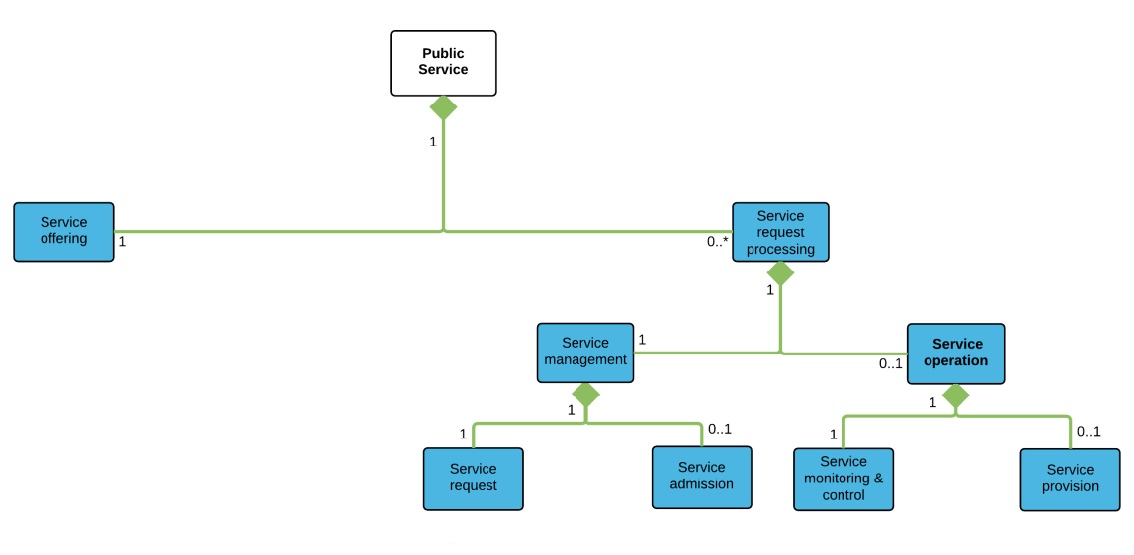}
    \caption{Offering and provision of a public service. Source: \cite{ref_proc2}}
    \label{fig:public_service}
\end{figure}

A public service consists of one activity of offering a given service and many activities of processing a request to use a given service - one for each request to use the service \cite{ref_proc2}. In addition, the request to use the service triggers activities related to the management of the given service, such as activities that process the request to use the service and decide whether to enable the given service to the requester, as well as service provision activities and service provision control and monitoring activities \cite{ref_proc2}. For each public service, one and only one uniquely identifiable service offering activity is required, whereas there may be zero or more activities processing service usage requests. Therefore, the service provision activity indicator also includes the service as a whole, which may include several activities of providing a given service but does not identify with any of them \cite{ref_proc2}. Activities concerned with offering and provision of public services are represented in Fig. \ref{fig:public_service} that originates in \cite{ref_proc2}.

\newpage
According to the Regulation of the Government of Estonia ``Basics of Service Organization and Information Management'' \cite{ref_url1}, a proactive service is a direct public service that the institution provides on its own initiative, on the basis of the assumed will of individuals and on the basis of the data in the databases belonging to the state information system. Proactive service is provided automatically or with the person's consent. According to the same regulation \cite{ref_url1}, event service is a direct public service that several institutions provide jointly so that a person can fulfill all obligations and use all rights that arise due to one event or situation. The event service aggregates several services related to the same life event into one service for the user. 

In order to include proactive and life event services, it is necessary to change the definition of public service provided in \cite{ref_proc2}. Namely, in the case of a proactive event service, there may be a lack of a direct desire to use the service, which is why we replaced the concept of service request processing introduced in \cite{ref_proc2} and represented in Fig. \ref{fig:public_service} with the concept of service initialization processing shown in Fig. \ref{fig:metamodel}. According to Fig. \ref{fig:metamodel}, a life event triggers zero or more service initialization processing activities, each of which includes activities related to the management and delivery of the corresponding service(s). The service management activity includes the service admission activity, which typically has to comply with specific rules. In case of life event services, such rules can be seen as business rules of the type \textit{reaction rules} \cite{ref_proc3,ref_proc5} that define the conditions under which the public service is triggered by the given life event. Such conditions may include checking if the person has opted out from the proactive electronic service format \cite{ref_proc4}, in which case the service is not provided before an explicit request.

The notion of life event is also included by the Core Public Service Vocabulary Application Profile (CPSV-AP) by the European Commission \cite{ref_url5}, which is a simplified, reusable and extendable data model that describes the main attributes of services offered by public administrations. The purpose of the CPSV-AP is twofold: to standardise catalogues of public services offered in the EU countries and to simplify and standardise data exchange for applying digital public services within EU. The Life Event class of the CPSV-AP represents an important event or situation in a citizen's life where public services are required. In the context of the CPSV-AP, the Life Event class only represents an event to which a public service is related. For example, a couple becoming engaged is not a CPSV-AP Life Event, while getting married is, since only the latter has any relevance to public services \cite{ref_url5}. In this sense, the Life Event class of the CPSV-AP is narrower than the Life Event concept included by the ontological metamodel depicted in Figure \ref{fig:metamodel}, which is more similar to the notion of data event that was introduced above in this section.

\section{Conclusions}
\label{conclusions}

Estonia is the first member state of the European Union that has legislatively defined proactive life event services. Surely, other European Union member states and the European Commission will soon follow Estonia in this regard. To avoid terminological confusion, provision of proactive life event services should be based on an ontological metamodel. This is essential for proper implementation of hardware and software architectures of e-government, but also for standardization and interoperability within EU.

In this paper, we propose the metamodel for proactive life event services, which is based on \cite{ref_article1} and extends \cite{ref_proc2}. In the future, we plan to delve more deeply into the topic of admission rules of public services that ultimately define the level and type of proactivity, which have been categorized in \cite{ref_proc6,ref_article2}.

\bibliographystyle{splncs04}
%\bibliography{mybibliography}
%

\end{document}